\documentclass[conference]{IEEEtran}
\IEEEoverridecommandlockouts

\usepackage{amsmath,amssymb,amsfonts}
\usepackage{algorithmic}
\usepackage{graphicx}
\usepackage{textcomp}
\usepackage{xcolor}
\usepackage{booktabs}
\usepackage{pifont}
\usepackage{easy-todo}
\usepackage{tabularx}
\usepackage{hyperref}
\usepackage{url}
\usepackage{fancyhdr}

\usepackage[backend=biber,
            hyperref=true,
            url=false,
            isbn=false,
            doi=false,
            backref=false,
            style=ieee,
            citestyle=numeric-comp,
            sorting=none,
            block=none]{biblatex}

\begin{document}
\title{ARISTO Hand: Sensing-Driven Distal Hyperextension for Fine-Grained Manipulation\\
\thanks{
This work was supported in part by Sony Group Corporation.

\textsuperscript{1} Authors are with Human Centered Robotics Lab at The University of Texas at Austin, TX, USA. (email: akim.jinsoo@utexas.edu)

\textsuperscript{2} Authors are with Sony Group Corporation at Tokyo, Japan.
}}


\author{
\IEEEauthorblockN{
Aaron Kim\textsuperscript{1},
Dong Ho Kang\textsuperscript{1},
Mark Helwig\textsuperscript{1},
Mingyo Seo\textsuperscript{1},
Kazuto Yokoyama\textsuperscript{2},
Tetsuya Narita\textsuperscript{2},
Luis Sentis\textsuperscript{1}
}
}

\maketitle

\pagestyle{fancy}
\fancyhf{}
\fancyhead[L]{\scriptsize IEEE/ASME AIM 2026, Preprint Version, Accepted May 2026}
\fancyhead[R]{\scriptsize Kim \textit{et al.}: ARISTO Hand}
\renewcommand{\headrulewidth}{0pt}
\thispagestyle{fancy}

\begin{abstract}
Manipulating thin objects requires precise contact geometry and reliable force perception, yet many anthropomorphic robotic hands lack the mechanical and sensing capabilities needed for such interactions. We present the ARISTO Hand, a tendon-driven robotic hand that integrates active distal hyperextension with a hybrid fingertip-sensing architecture that combines a rigid, nail-mounted force-torque sensor and a soft capacitive tactile array. Active hyperextension enables controlled fingertip engagement beyond the kinematic limits of standard flexion, increasing pull-out force by $2.76\times$ for object thicknesses of $1$-$20$\,mm while preserving the nominal grasp capability. The rigid nail-mounted sensor provides reliable force measurements during edge contacts, where the sensitivity of proprioceptive force estimation degrades as the contact geometry approaches kinematic singularities. We validate the proposed architecture through quantitative force characterization and a multi-stage SD card extraction and insertion task. Video and supplementary materials are available at:
\url{https://aristohand.github.io}.
\end{abstract}



\section{Introduction}
As robots evolve to support increasingly complex tasks, their value is defined by the capacity to interact with the physical world, placing manipulation at the forefront of their capabilities. Although recent advances in hardware, control, and learning have enabled progress in many manipulation tasks \cite{shaw_leap_2023, shaw_leap_2025, christoph_orca_2025, yuan_development_2026}, small, thin, and fragile objects remain particularly challenging. Everyday household and workplace tasks expose this gap, from peeling tape to lifting a thin plastic tab, where hands must align their fingertips to a surface, detect subtle changes in contact, and securely interface with small edges. These tasks demand precise surface contact, sensitivity to micro-scale forces, and consistent fingertip orientation during interaction \cite{do_densetact-mini_2024, miyama_design_2022}. As a result, the inability to reliably perform these interactions remains a fundamental barrier to deploying robots within everyday human environments.

\begin{figure}[t]
\centering
\includegraphics[width=1.0\linewidth]{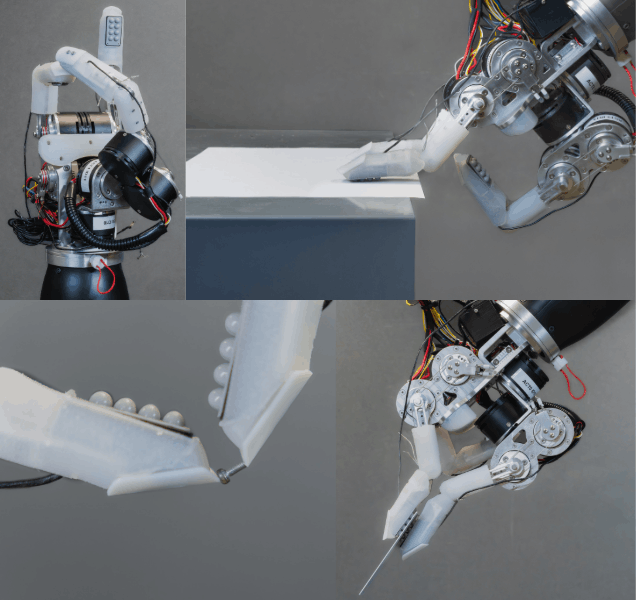}
\caption{\textbf{Overview of the ARISTO Hand and representative fine-grained manipulation behaviors enabled by active distal hyperextension and integrated rigid-soft fingertip sensing.}
(\textit{Top-left}) The ARISTO Hand mounted on a 7-DoF robotic arm.
(\textit{Top-right}) Distal joint hyperextension enabling the fingertip pad to align parallel to a planar surface.
(\textit{Bottom-left}) Close-up of the rigid nail-mounted force-torque sensor during fine-grained interaction.
(\textit{Bottom-right}) Hyperextension enabling stable grasp acquisition on a thin object through surface-aligned fingertip contact.}
\label{fig:aristo-examples}
\end{figure}

Importantly, no single sensing or mechanical strategy is sufficient for these interactions; real-world thin-object tasks require a coordinated combination of geometric adaptability and high-fidelity sensing. Despite substantial progress in dexterous hand design, existing approaches often address only a narrow portion of these requirements. On the sensing side, while resolution and packaging have improved \cite{taylor_fully_2024, kim_anthropomorphic_2023}, most hands rely on a single sensing modality. This reliance limits the ability to distinguish distinct contact modes (e.g., fingernail vs. pad) or perceive incipient shear \cite{ma_gellink_2024, romero_eyesight_2024}. On the mechanical side, designs prioritizing anthropomorphic kinematics often feature smooth or uniformly compliant fingertips \cite{nikafrooz_single-actuated_2021, sung_snu-avatar_2023}. While effective for grasping palm-sized objects, these geometries struggle to securely interface with small edges or maintain stable contact against flat surfaces \cite{zhou_3d_2021, baker_star2_2023, liu_passively_2024}. Together, these design choices prioritize robustness and generality over precise distal interaction, leaving small-object manipulation insufficiently supported.

Reliable small-object manipulation, therefore, demands two complementary capabilities. First, a multi-layered sensing suite in which distinct feedback from a rigid fingernail and soft fingertip is complemented by proprioceptive torque estimation, ensuring contact observability across varied interaction modes \cite{kang_plato_2026}. Second, purposeful mechanical adaptation is required to align with flat surfaces and access constrained regions. Crucially, these capabilities are interdependent: sensing without geometric control cannot ensure proper contact conditions, while geometric control without detailed feedback is unsafe. Together, they define the requirements for a hand capable of actively shaping contact at the fingertip when interacting with thin, fragile objects.

To realize this, we introduce the ARISTO (\underline{A}nthropomorphic, \underline{R}obotic, \underline{I}ntegrated-\underline{S}ensing, \underline{T}endon-\underline{O}perated) Hand as a study platform that unifies advanced fingertip sensing with purposeful mechanical adaptation for reliable fine-grained manipulation. Mechanically, the hand incorporates independently actuated distal hyperextension, enabling systematic investigation of how active fingertip alignment with flat surfaces influences contact stability. Sensorially, it integrates a rigid sensorized fingernail with a high-resolution tactile pad, allowing examination of how distinct sensing modalities contribute to force awareness and contact discrimination along thin edges.

The primary contributions of this work are:
\begin{itemize}
    \item A tendon-driven distal hyperextension mechanism that enables active fingertip alignment for stable thin-object manipulation.
    \item An integrated rigid-soft fingertip sensing architecture combining nail-mounted force-torque sensing for edge interaction with compliant tactile slip monitoring for stable grasp acquisition.
    \item Experimental validation that fingertip geometry strongly influences both force transmission and proprioceptive force observability during fine-grained manipulation.
\end{itemize}
\section{Related Work}

The manipulation of small and thin objects imposes two distinct requirements often treated separately in the literature: the mechanical capacity for precise distal reorientation to align with flat surfaces \cite{kang_three-fingered_2021}, and the sensory capacity to discriminate between thin edges and distributed contacts \cite{do_densetact-mini_2024}. While numerous anthropomorphic hands have been developed, existing platforms typically optimize for either robust power grasping or soft tactile exploration, leaving a functional gap for tasks requiring both rigid edge engagement and active fingertip adaptation.

Our previous work, the PLATO Hand, explored how rigid-compliant fingertip geometry and force-transparent linkage transmission improve manipulation through deformation-driven contact stabilization \cite{kang_plato_2026}. However, the linkage-driven architecture limited independent distal reconfiguration and employed mechanically coupled sensing. In contrast, ARISTO investigates actively controlled distal hyperextension and mechanically decoupled rigid-soft sensing to study how fingertip geometry and sensing modality jointly influence thin-object manipulation.

\subsection{Geometric Adaptation and Surface Alignment}

Stable acquisition of thin objects requires the fingertip to actively align with the environment \cite{kang_three-fingered_2021}. Existing anthropomorphic hands generally approach this through either tendon-driven or linkage-based architectures, yet both face limitations in controlling distal interactions.

Tendon-driven systems, including ORCA, LEAP V2, and DexHand~021, prioritize compliance and impact robustness \cite{christoph_orca_2025,shaw_leap_2025,yuan_development_2026}. However, their compact anthropomorphic form factors often rely on underactuation and complex tendon routing, introducing joint coupling and hysteresis that limit independent distal adjustment \cite{wu_adaptive_2025,nikafrooz_single-actuated_2021,fajardo_guaranteed_2024}.

Conversely, linkage-driven designs such as OPENGRASP-LITE, the Everyday Finger, and the ILDA Hand provide high stiffness and analytically tractable kinematics, enabling predictable mappings between actuator input, fingertip motion, and force transmission \cite{gros_opengrasp-lite_2024,ornelas_everyday_2024,kim_integrated_2021}. Consequently, a key challenge remains in combining the stiffness and predictability of linkage mechanisms with the independent configurability of tendon-driven systems.

\subsection{Sensing Fidelity at the Contact Interface}

Reliable manipulation additionally requires distinguishing between localized edge contact and distributed surface pressure. Existing sensing approaches often trade material softness for geometric precision, as compliant sensing surfaces can blunt thin contact features, attenuate force transients, and limit directed edge interaction \cite{do_densetact-mini_2024,christoph_orca_2025,junge_spatially_2025}.

Vision-based tactile sensors such as GelLink and DenseTact-Mini provide rich contact information, while ROMEO extends tactile sensing to the palm for whole-hand interaction \cite{ma_gellink_2024,do_densetact-mini_2024,liu_passively_2024}. However, the soft elastomeric skins required for these sensors can deform substantially under load, reducing the sharp localized geometry needed for prying or scraping thin edges \cite{taylor_fully_2024,do_densetact-mini_2024}.

Alternatively, proprioceptive and quasi-direct-drive hands, such as the EyeSight Hand, offer high transparency in force control without fragile tactile skins \cite{romero_eyesight_2024}. However, recent studies show that internal sensing alone is insufficient for reliably interpreting interaction forces in compliant hands \cite{junge_spatially_2025}. In extended poking or prying configurations, external forces can produce weak or ambiguous motor-side signatures, particularly in underactuated systems with mechanical coupling and limited actuation authority \cite{ma_gellink_2024}. Consequently, relying solely on either soft tactile sensing or internal motor estimation leaves critical gaps in force observability during fine-grained manipulation.
\section{Design}

\begin{figure*}[t]
    \centering
    \vspace{2mm}
    \includegraphics[width=1.0\textwidth]{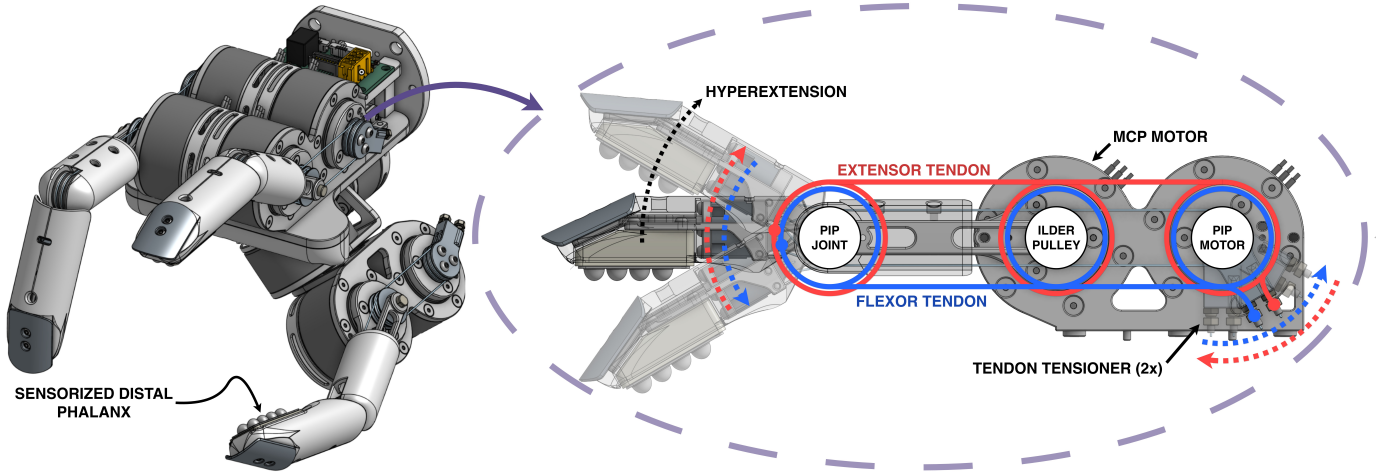}
    \vspace{-2mm}
    \caption{(\textit{Left}) CAD model of the ARISTO Hand. (\textit{Right}) Antagonistic tendon routing diagram. Flexor and extensor tendons are fully wrapped around all pulleys, ensuring a configuration-independent moment arm and constant-tension transmission. This routing supports linear torque generation and active distal hyperextension.}
    \label{fig:cad}
\end{figure*}

\subsection{Mechanical Architecture}

    The ARISTO Hand is designed to enable controlled distal contact, predictable force transmission, and stable interaction across flexion and hyperextension. Each finger employs a compact two-joint architecture driven by a fully wrapped antagonistic tendon transmission (Fig.~\ref{fig:cad}).

\subsubsection{Rigid, Configuration-Independent Transmission}

Each finger employs a fully wrapped antagonistic tendon routing that preserves a constant effective moment arm $r$ across the joint range of motion (Fig.~\ref{fig:cad}). Because the tendon lift-off point remains tangential to the pulley throughout rotation, the joint torque follows

\[
\tau_j = r \left(f_{\mathrm{flex}} - f_{\mathrm{ext}}\right),
\]

where $f_{\mathrm{flex}}$ and $f_{\mathrm{ext}}$ are the flexor and extensor tendon tensions. Since $r$ remains approximately constant, the actuator effort maps linearly to joint torque independent of configuration.

This linear torque relationship is critical for controlled distal interaction. In conventional tendon-driven systems, configuration-dependent moment arms introduce nonlinear torque mapping, complicating force regulation and reducing transparency. By maintaining a constant moment arm, ARISTO delivers consistent torque across flexion and hyperextension, enabling stable force application during edge engagement and thin-object manipulation.

The routing is preloaded with braided stainless-steel tendons, tensioned via a vented screw mechanism that maintains constant tension, eliminating backlash without introducing passive compliance from cable stretch. As a result, actuator effort remains directly coupled to joint torque, supporting reliable proprioceptive load estimation and high-fidelity transmission of fingertip contact forces.

\subsubsection{Active Distal Hyperextension}

The distal joint supports active hyperextension beyond neutral alignment. This capability allows the fingertip pad to flatten against planar surfaces and rotate past conventional flexion-only limits.

Hyperextension increases effective contact area during wedging, sliding, and thin-object engagement. It also maintains more favorable joint configurations during edge interaction, reducing the need for excessive proximal flexion and improving mechanical advantage.

\subsubsection{Fingernail Geometry: Angle and Curvature}

The rigid fingernail is angled downward relative to the distal link rather than parallel to it, allowing the nail tip to engage edges and slide beneath thin objects without requiring excessive PIP flexion. By pre-orienting the contact surface, the finger can perform prying and undercutting motions while maintaining favorable joint configurations.
In addition, the nail surface is slightly curved rather than flat. This curvature centralizes small objects during pinch grasps and reduces lateral slip, guiding narrow or cylindrical objects toward the fingertip centerline and improving grasp stability.

\subsection{Torque Transparency and Proprioceptive Estimation}

ARISTO employs quasi-direct-drive (QDD) actuation with a low transmission reduction to preserve torque transparency and backdrivability at the distal joint. Compared to high-ratio transmissions, this configuration minimizes reflected inertia and transmission friction.

The low reduction ratio enables reliable joint torque estimation from motor current and supports smooth transitions between compliant and high-stiffness behavior during constrained interaction. While joint torque provides accurate actuator-level load estimation, it does not directly encode the distal contact wrench. External contact forces remain configuration-dependent; therefore, fine manipulation tasks rely primarily on distal sensing modalities.

\subsection{Integrated Fingertip Sensing}

\begin{figure}[t]
    \centering
    \vspace{2mm}
    \includegraphics[width=1\linewidth]{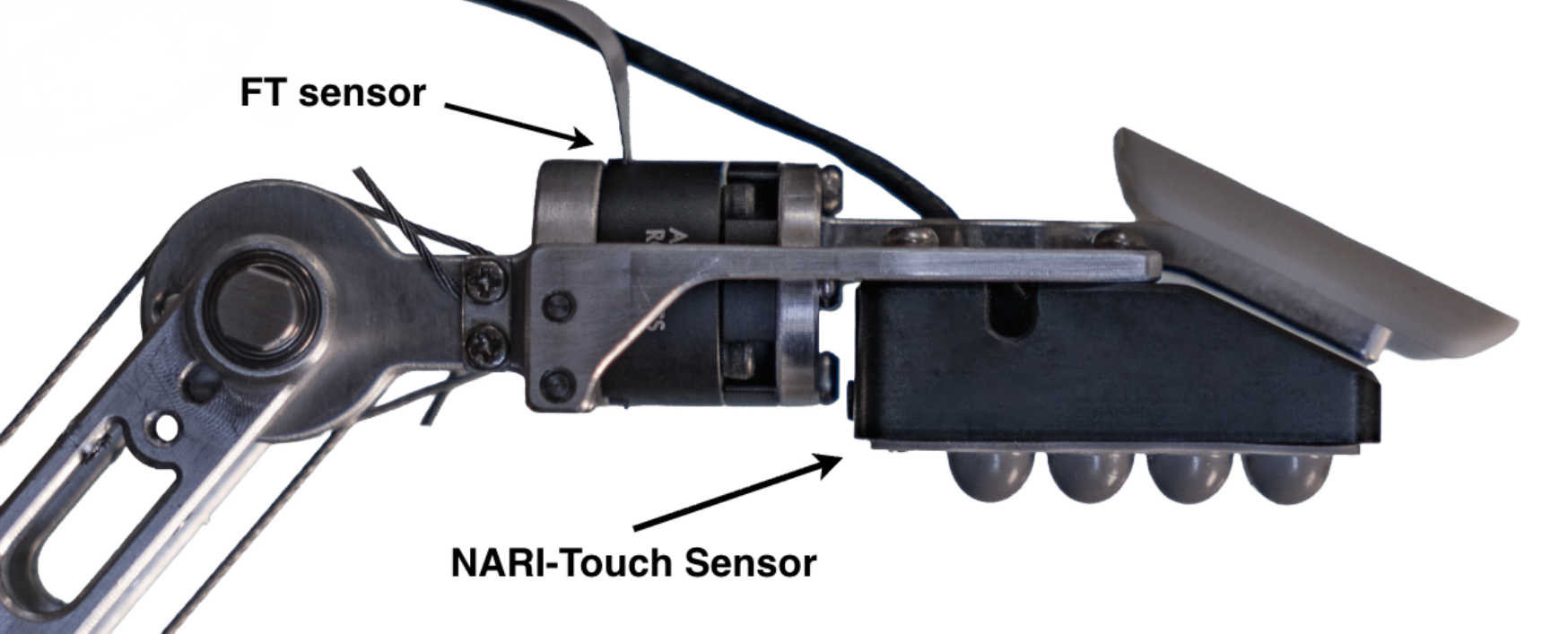}
    \caption{\textbf{Integrated fingertip sensing.} A rigid nail-mounted force–torque sensor enables edge-sensitive interaction, while a mechanically isolated soft capacitive array monitors distributed surface contact.}
    \label{fig:fingertip_sensing}
\end{figure}

The fingertip integrates two mechanically decoupled sensing modalities to balance edge sensitivity and compliant contact (Fig.~\ref{fig:fingertip_sensing}).

\subsubsection{Rigid Nail and Soft Pad}

A rigid dorsal fingernail is mounted to a miniature six-axis force–torque sensor (AIDIN Robotics \cite{noauthor_adn-aft20-d15_nodate}), directly measuring the contact wrench $\mathbf{w}_{\mathrm{nail}} \in \mathbb{R}^6$ at the nail interface. The rigid coupling preserves high-frequency force transients and localized edge interactions that would otherwise be attenuated by compliant materials.

The ventral surface incorporates a NARI-Touch $2 \times 4$ capacitive tactile array \cite{narita_theoretical_2020} embedded within an elastomeric pad, producing a distributed pressure output $\mathbf{z}_{\mathrm{tac}} \in \mathbb{R}^8$ for monitoring surface contact and incipient slip.

\subsubsection{Modal Complementarity and Redundancy}

Mechanical isolation between the rigid nail and soft pad prevents cross-modal interference while enabling task-dependent sensing. The nail-mounted force–torque sensor supports geometry-constrained tasks such as edge scraping, levering, and recessed interaction. The soft tactile array supports frictional surface acquisition and stable pinch grasps through distributed pressure feedback.

In configurations where distal tactile contact is limited or partially occluded, proprioceptive torque estimation provides a secondary load signal. Although joint torque does not directly encode the full distal wrench, it enables consistent monitoring of applied force during constrained interaction. Together, the rigid nail sensor, compliant tactile pad, and torque transparency form a multimodal sensing architecture aligned with controlled distal contact.

\section{Experiments}

We evaluate whether the mechanical and sensing principles of the ARISTO Hand support reliable manipulation of small, thin objects during contact-rich fingertip interactions. These interaction regimes require precise control of contact geometry and force, rather than conventional force closure in free space.

All experiments focus on fine-grained interactions involving thin objects, edge contact, or geometrically constrained access, such as recessed slots, planar contact alignment, and guided pull-out motions. Fingertip sensing data were recorded at 200~Hz and include a six-axis force-torque (FT) sensor mounted beneath the rigid fingernail and a $2 \times 4$ capacitive tactile array embedded in the fingertip pad. Ground-truth force measurements were collected using external sensors where applicable.

The experiments are organized as follows. Section~IV-A establishes baseline joint-level performance and validates the force transparency of tendon-driven transmission. Section~IV-B characterizes the fidelity and limitations of the fingertip sensing modalities. Section~IV-C presents task-level demonstrations showing how multimodal fingertip sensing and actively commanded distal hyperextension enable robust thin-object manipulation and constrained edge interaction.

\subsection{Joint-Level Characterization}

We characterized the ARISTO Hand using the NIST Performance Metrics for Robotic Hands \cite{falco_performance_2018}, with a Bota Systems six-axis force–torque sensor \cite{noauthor_rokubimedusa_nodate} used as ground truth. Measured metrics are summarized in Table~\ref{tab:performance_metrics}.

\begin{table}[h]
\centering
\caption{NIST Performance Metrics for the ARISTO Hand}
\label{tab:performance_metrics}
\begin{tabular}{@{}lll@{}}
\toprule
\textbf{Metric} & \textbf{Value} & \textbf{95\% CI} \\ \midrule
Fingertip Force (Continuous) & 4.88 N & --- \\
Grasp Strength (80mm cylinder) & 17.81 N & [17.80, 17.82] N \\
Grasp Cycle Time & 448.44 ms & [444.80, 452.08] ms \\ \bottomrule
\end{tabular}
\end{table}

The measured fingertip and grasp forces are lower than those of high-reduction commercial hands, reflecting the use of quasi-direct-drive (QDD) actuation with an 8:1 gear ratio and a three-finger configuration. Rather than maximizing force amplification, the system prioritizes torque transparency and dynamic responsiveness. The measured grasp cycle time of 0.448~s reflects the high actuation bandwidth enabled by the low-reduction transmission.

\begin{table}[h]
\centering
\footnotesize
\caption{Hand Performance Comparison (NIST Metrics)}
\label{tab:hand_comparison}
\begin{tabular}{lccc}
\toprule
\textbf{Hand} & \textbf{Finger (N)} & \textbf{Grasp (N)} & \textbf{Time (s)} \\
\midrule
\textbf{ARISTO} & \textbf{4.88} & \textbf{17.81} & \textbf{0.448} \\
Schunk DH & 22.11 & 76.11 & --- \\
Robotiq 3F & 39.65 & 92.97 & 3.82 \\
\bottomrule
\end{tabular}
\end{table}

Table~\ref{tab:hand_comparison} contextualizes performance relative to representative hands evaluated under the same NIST protocols. While commercial grippers achieve substantially higher grasp forces, they operate with slower cycle times. ARISTO instead emphasizes responsive, torque-transparent interaction suitable for contact-sensitive manipulation.

\subsubsection{Transmission Linearity Validation}

To evaluate transmission linearity, we characterized the relationship between motor effort and externally applied force while isolating the proximal interphalangeal (PIP) joint. The metacarpophalangeal (MCP) joint was mechanically fixed, and the PIP joint was commanded to maintain a fixed fingertip pose in the world frame while external forces were applied using a calibrated force–torque sensor.

Across MCP configurations spanning $-60^\circ$ to $+60^\circ$, motor effort exhibited a strong linear relationship with applied force. The mean coefficient of determination was $R^2 = 0.982 \pm 0.007$, with a mean normalized root-mean-square error of 4.41\%. The fitted force–effort slope varied by only 2.14\% across configurations. A global fit across all configurations yielded

\[
F = 20.13u + 0.71,
\]

demonstrating consistent effort-to-force mapping through both flexion and hyperextension.

\begin{figure}[t]
    \centering
    \includegraphics[width=\linewidth]{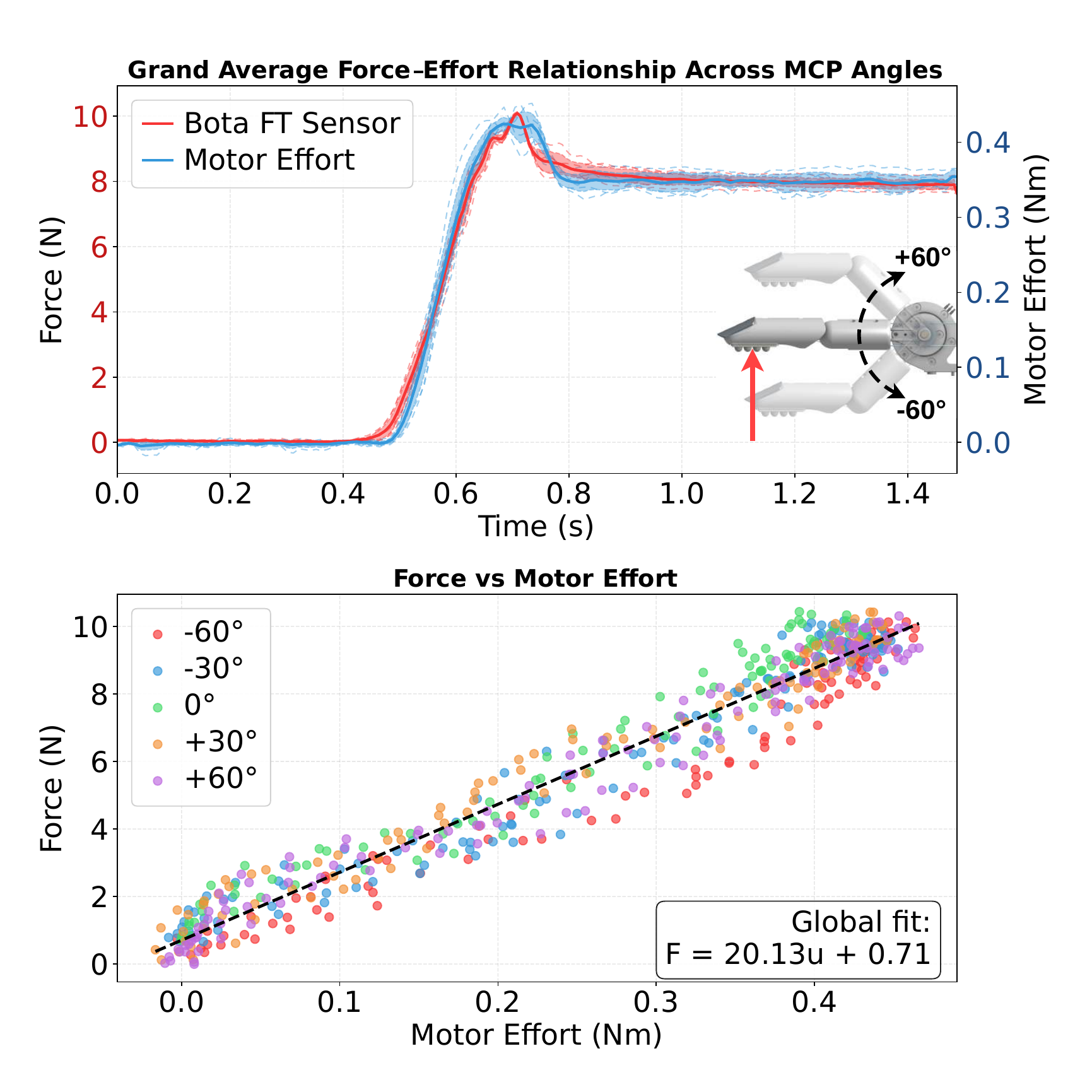}
    \caption{\textbf{Transmission linearity validation.}
    \textit{(Top)} Grand average force–effort relationship across MCP configurations with the PIP joint isolated; shaded region denotes the 95\% confidence interval.
    \textit{(Bottom)} Motor effort versus externally applied force for individual MCP configurations ($-60^\circ$ to $+60^\circ$). Relationships remain approximately linear ($R^2 = 0.982 \pm 0.007$), with a slope coefficient of variation of 2.14\%.}
    \label{fig:force_validation}
\end{figure}

\subsubsection{Hyperextension Efficacy for Thin Object Manipulation}

To quantify the benefit of actively controlled distal hyperextension for stabilizing contact on thin objects, we conducted a pull-out force ablation study across object thicknesses ranging from approximately 0.75~mm (credit-card thickness) to 100~mm. All experiments used a symmetric two-finger pinch grasp. Pull-out tests were performed using an Instron 3367 tensile testing machine \cite{noauthor_instron_nodate-1}, which applied a constant-velocity vertical displacement, while forces were measured with an Instron 2530-30kN load cell \cite{noauthor_instron_nodate}. In each trial, the grasped block was rigidly mounted and pulled vertically until slip or loss of contact occurred, and the maximum sustainable pull-out force was recorded.

This study does not use sensor feedback, as the goal is to isolate the kinematic effects of distal hyperextension without confounding effects from sensing or control. Force levels were intentionally limited to avoid actuator saturation, thereby enabling consistent motor behavior across many repeated trials. Consequently, the reported average pull-out force of approximately 3.5~N for thick objects does not represent the system’s maximum force capability. When measuring maximum pull-out force without constraining motor gains, the system achieved a peak force of 10.13~N.

Two kinematic configurations were evaluated. The standard flexion configuration, representative of flexion-only fingertip kinematics commonly used in anthropomorphic hands, restricted the distal joint to positive angles, resulting in edge or curved pad contact. In the hyperextension configuration, the distal joint was actively commanded to hyperextend, allowing the fingertip pad to flatten against the object surface and increase contact area.

\begin{figure}[t]
    \centering
    \includegraphics[width=1\linewidth]{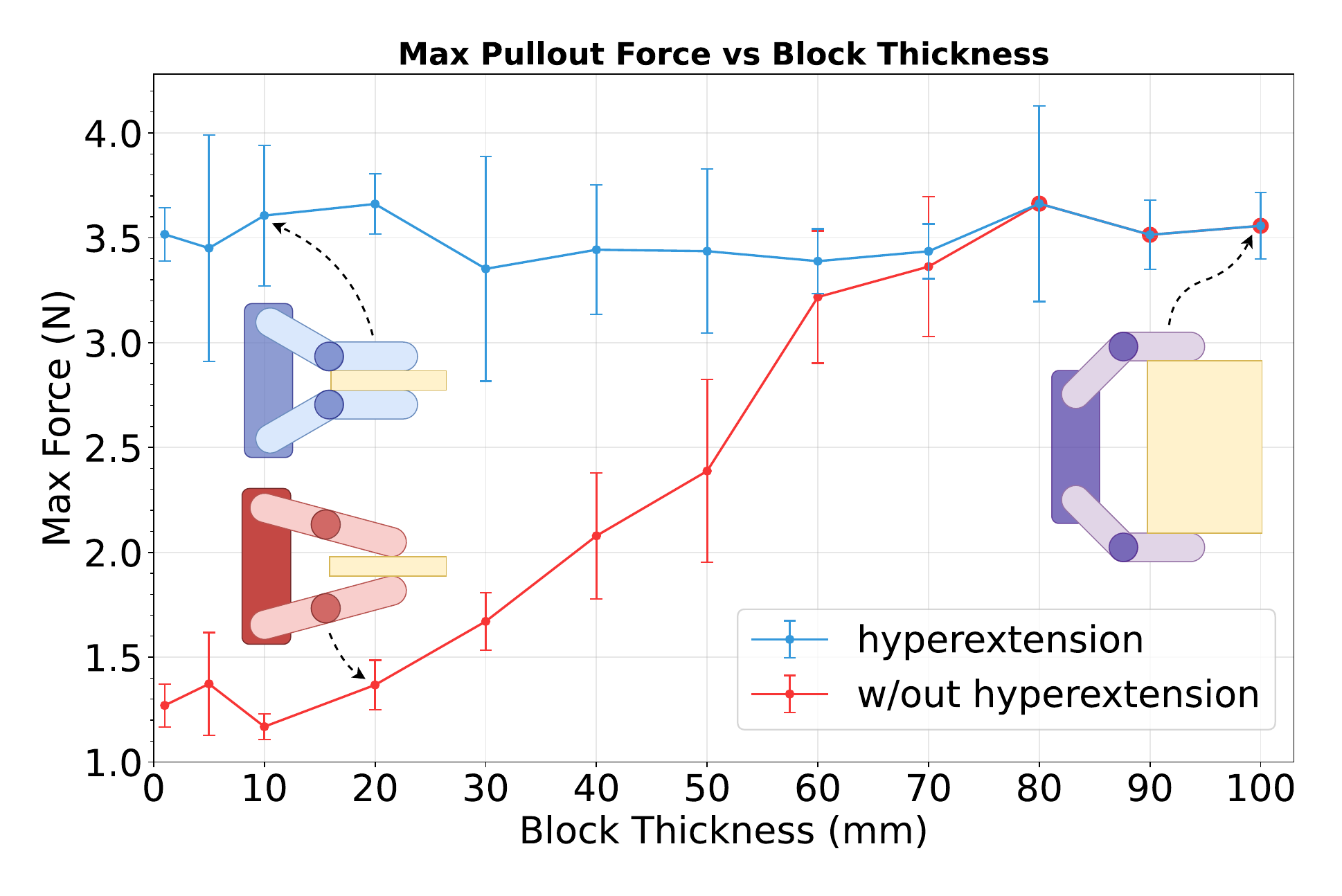}
    \caption{\textbf{Effect of distal hyperextension on thin-object pull-out force.}
Maximum pull-out force versus block thickness for hyperextension and standard flexion. Points show trial means with error bars indicating $\pm$ one standard deviation. Hyperextension preserves higher and more consistent pull-out forces for thin objects ($\leq 30$ mm), while both configurations converge for thicker objects ($\geq 70$ mm).}
    \label{fig:hyperextension_pullout}
\end{figure}

As shown in Fig.~\ref{fig:hyperextension_pullout}, both configurations achieved comparable pull-out forces for thick objects ($\geq 70$~mm), with average forces of approximately 3.5~N, as standard flexion alone is sufficient to achieve near-parallel surface contact. Performance converges between 60-70~mm. Below this threshold, standard flexion increasingly forces edge contact, leading to a rapid reduction in achievable pull-out force. In contrast, distal hyperextension maintains parallel surface contact by rotating the fingertip pad opposite the PIP joint direction. For thin objects (1-20~mm), hyperextension produced a mean pull-out force 2.76$\times$ higher than standard flexion. Normalized by the thick-object baseline, hyperextension retained at least 98\% of available pull-out force down to credit-card thickness, whereas standard flexion retained only 34-40\%.

\subsection{Sensing Fidelity and Modality Validation}

We evaluated the necessity of distal sensing by comparing the performance of the nail-mounted FT sensor against proprioceptive motor effort in two critical scenarios: (1) edge contact detection at low velocities and (2) geometry-dependent normal force detection.

\subsubsection{Low-Velocity Edge Contact Detection}

To simulate the precise approach required to engage a thin edge, the finger was commanded to press the fingernail tip against a rigid edge mounted on a ground-truth Bota FT sensor. We evaluated each modality's ability to distinguish the moment of contact from free-motion transmission dynamics.

\begin{figure}[t]
    \centering
    \includegraphics[width=\linewidth]{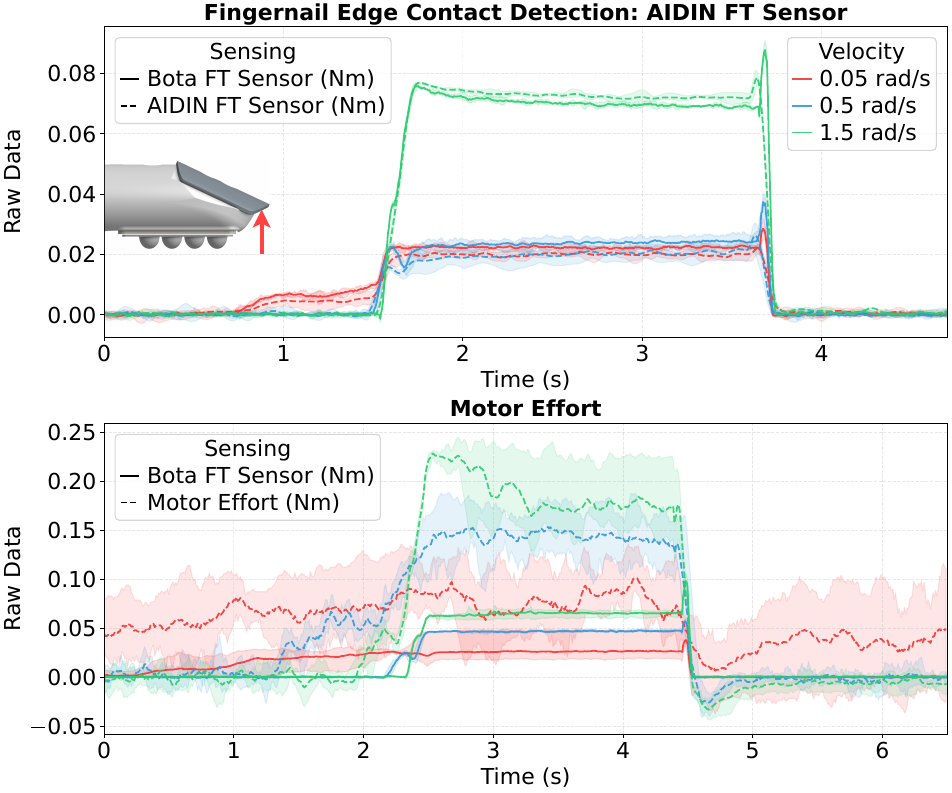}
    \caption{\textbf{Edge contact at varying approach velocities.}
    \textit{(Top)} The AIDIN FT sensor exhibits clear, repeatable contact steps independent of velocity, closely matching ground truth.
    \textit{(Bottom)} Motor effort is noisy and velocity-dependent; at low speeds (0.05~rad/s), stiction obscures contact events, demonstrating the unreliability of proprioception alone for precise edge engagement.}
    \label{fig:fingernail_edge}
\end{figure}

As shown in Fig.~\ref{fig:fingernail_edge}, the AIDIN FT sensor provided clean, step-like contact detection across a range of approach velocities. At lower speeds (0.05 and 0.5~rad/s), where precision manipulation typically occurs, the FT sensor maintained a high signal-to-noise ratio with a pre-contact noise floor of only $\sigma = 0.0004$~Nm. In contrast, motor effort signals were inherently noisy despite using the same underlying controller. At 0.05~rad/s, stiction elevated the baseline motor torque to approximately $0.06$~Nm, completely obscuring the contact impulses visible in the ground-truth force data and preventing the use of a sensitive detection threshold without false positives.

While motor effort baselines varied significantly with approach velocity, the FT sensor response remained consistent, except at the highest speed (1.5~rad/s). The transient overshoot observed at this velocity is likely attributable to controller latency and the finger’s inertia, which limit how rapidly motor commands can stop motion at higher speeds. Importantly, even under these conditions, the FT sensor continued to exhibit a distinct contact signature, whereas motor effort remained dominated by velocity-dependent transmission effects. These results confirm that distal sensing is required to decouple contact forces from transmission friction during the delicate approach and edge-engagement phases.

\subsubsection{Geometry-Invariant Force Detection}

A critical limitation of proprioceptive sensing in extended fingers is the geometric singularity that occurs as the finger straightens. To quantify this, we commanded the finger to apply a normal force against a flat plate at varying MCP joint angles ($\theta \in [15^\circ, 60^\circ]$) while maintaining the PIP joint at $0^\circ$ in the world frame. We calibrated both modalities to target an approximate ground-truth force of 3.1~N.

\begin{figure}[t]
    \centering
    \includegraphics[width=\linewidth]{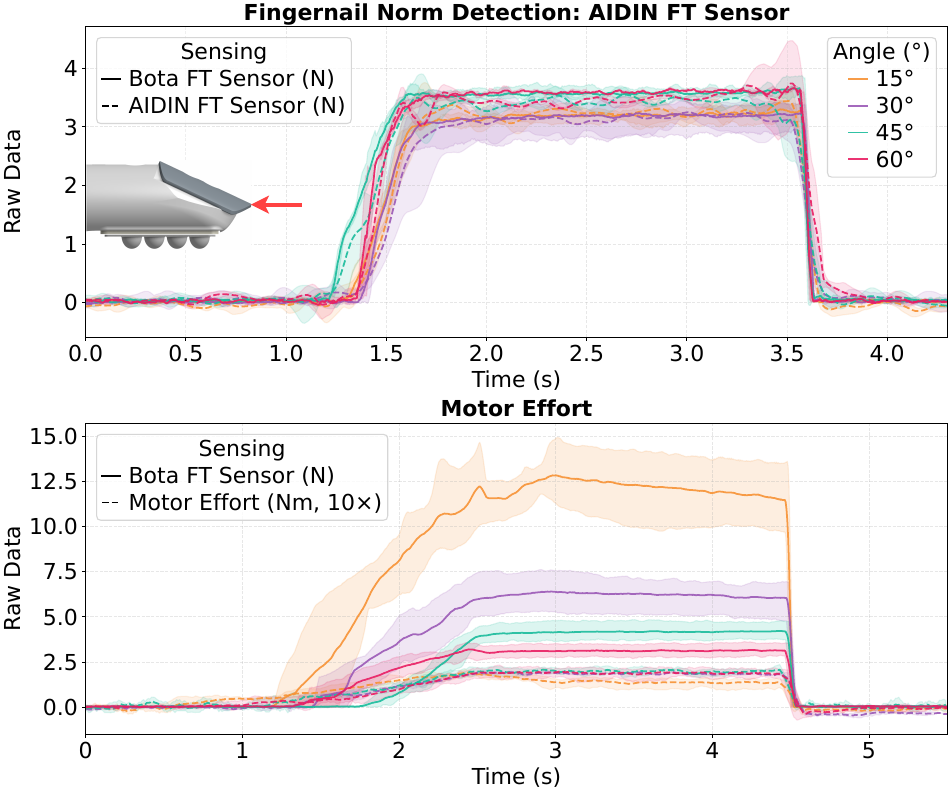}
    \caption{\textbf{Geometry-invariance of distal sensing versus proprioception.}
Normal force tracking on a flat surface across incidence angles of $15^\circ$--$60^\circ$. \textit{Top:} The distal FT sensor closely tracks ground truth across angles. \textit{Bottom:} Motor effort, scaled by $10\times$, degrades at shallow angles as the contact force approaches the joint axis, making normal force poorly observable from motor current. This highlights the need for distal sensing in edge-contact manipulation.}
    \label{fig:fingernail_norm}
\end{figure}

\begin{figure*}[t]
    \centering
    \vspace{2mm}
    \includegraphics[width=1.0\textwidth]{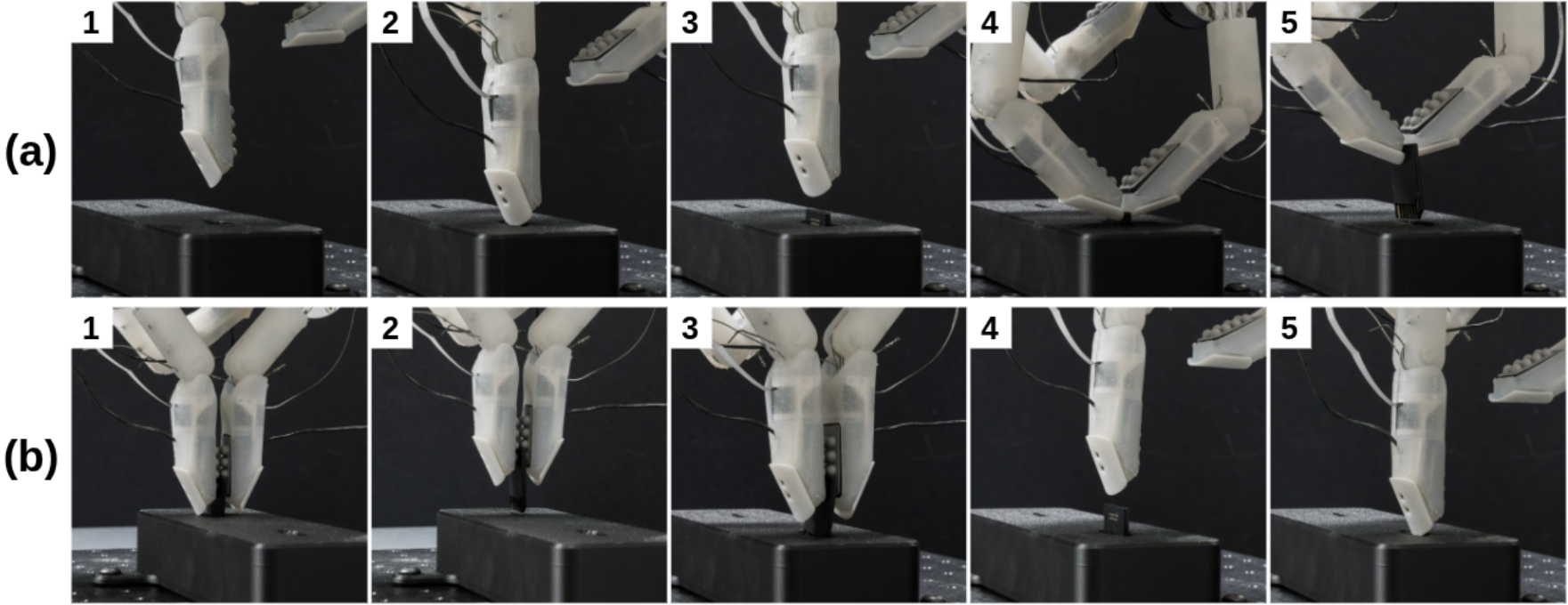}
    \vspace{-2mm}
    \caption{\textbf{End-to-end SD card manipulation demonstrating multimodal synergy.}
    \textbf{(a) Extraction:} (1-2) The rigid fingernail enters the recessed slot where soft pads cannot reach. (3) The FT sensor regulates the poking force to trigger the push-to-eject mechanism. (4-5) The nail pinches the card out for retrieval.
    \textbf{(b) Insertion:} (1) Active distal hyperextension aligns the fingertip pad parallel to the SD card for stable acquisition. (2-3) The soft tactile sensor enables a gentle, compliant grasp, facilitating alignment during insertion. (4-5) The rigid nail is used to push the card into the locked position.}
    \label{fig:sd_card}
\end{figure*}

Fig.~\ref{fig:fingernail_norm} illustrates the divergence in sensing reliability. The distal FT sensor demonstrated geometry-invariant performance, consistently triggering within $\pm 0.2$~N of the target force at all angles. Conversely, motor effort was highly sensitive to kinematics. As the finger approached a straighter configuration ($15^\circ$), the contact geometry approached a kinematic singularity, reducing the projection of the external force onto the joint torque space and causing the motor-side estimate to significantly underestimate the applied force. Consequently, when relying on motor effort to target 3.1~N at $15^\circ$, the system applied an actual force of $12.3$~N, an error of approximately $=300\%$. Since thin-object manipulation often requires these "poking" configurations, this result validates that transmission transparency alone is unsafe for fine manipulation. 

This result clarifies the role of quasi-direct-drive (QDD) actuation within ARISTO’s sensing architecture. QDD enables accurate estimation of joint torque from motor current due to its low friction and low reduction ratio. However, joint torque is distinct from distal contact wrench sensing. External forces depend on finger configuration and Jacobian mapping, which become ill-conditioned near singularities. Consequently, fine manipulation tasks such as slip detection, shear regulation, and poking primarily rely on distal sensing, whereas QDD provides reliable proprioceptive joint information.

\subsection{Application: SD Card Manipulation}

We validated the complete system (active hyperextension, rigid nail sensing, and soft tactile feedback) in a multi-stage SD card manipulation task. This task represents a contact-rich scenario involving a thin, fragile object constrained within a recessed push-to-eject slot, requiring precise force regulation and compliant alignment.

\subsubsection{Control Architecture}

All joints are controlled using joint-space impedance control with two operating modes: a compliant mode for free motion and a high-stiffness mode for constrained interaction. During approach, low stiffness improves robustness to unexpected collisions and reduces energy consumption. Upon contact detection, stiffness is increased to actuate constrained mechanisms such as the SD card eject spring. The low reflected inertia and backdrivability of the quasi-direct-drive actuation enable smooth transitions between these modes.

Contact detection at the rigid fingernail uses binary threshold logic based on the nail-mounted six-axis force-torque (FT) sensor. Thresholds were manually selected for each object and interaction condition to ensure reliable detection while remaining above the sensor noise floor. Upon contact detection, the controller transitions from a compliant mode to a high-stiffness interaction mode and holds the current joint position. 

In contrast, the soft tactile pad is used for continuous grasp monitoring rather than binary contact detection. During grasp acquisition and insertion, the NARI-Touch capacitive array monitors incipient slip, and the controller increases grip force when slip-related tactile changes are detected. Future work will investigate adaptive thresholding and closed-loop multimodal policies to eliminate manual tuning and generalize across object classes.

\subsubsection{Extraction via Rigid Sensing}

The extraction phase (Fig.~\ref{fig:sd_card}a) requires actuating the spring-loaded eject mechanism. This step is infeasible with standard soft fingertips, which are too compliant to depress the mechanism and too large to fit into the recessed slot. ARISTO utilizes the rigid fingernail to access the recess. In the first stage, the system uses the nail-mounted FT sensor to regulate the normal ``poking'' force (Section IV-B), ensuring the eject mechanism is triggered without over-stressing the card.

After ejection, the nominal strategy is to use the fingertip pad (NARI-Touch) to establish a stable pinch grasp and lift the card. However, in practice, the SD card is not always ejected far enough for the tactile pad, positioned further back on the fingertip, to make contact. In this geometrically occluded configuration, sensor redundancy is necessary: ARISTO continues to use the fingernail and monitors the torque components measured by the nail-mounted FT sensor to confirm sufficient applied load for engagement, while avoiding excessive force on the SD card.

Finally, the SD card is pulled out using the fingernails. While this contact does not provide direct shear/slip measurement at the fingertip pad, the nail-mounted FT sensor still provides meaningful force/torque feedback during extraction, which is preferable to executing the pull-out motion without any contact sensing.
  
\subsubsection{Insertion via Hyperextension and Tactile Feedback}

The insertion phase (Fig.~\ref{fig:sd_card}b) leverages the active hyperextension and tactile pad. To acquire the SD card from a preconfigured standing position, the finger hyperextends to align the tactile sensor with the card face, maximizing contact area for a stable pinch grasp. The NARI-Touch capacitive sensor provides a gentle grip, essential for handling sensitive electronics, while the soft pad's compliance passively compensates for minor misalignments during the ``peg-in-hole'' insertion process. Finally, the rigid nail is used again to push the SD card inwards; once we detect that the locking mechanism has been triggered (i.e., the card has bottomed out in the slot), the hand withdraws.

This sequence demonstrates that the mechanical and sensory features of ARISTO are complementary rather than redundant: the nail handles geometric constraints and rigid actuation, while the hyperextending pad handles stable acquisition and compliant alignment. Additionally, this task highlights the value of sensor redundancy in overcoming geometric design constraints. When the tactile sensor is geometrically unable to engage the card during early extraction, the distal fingernail sensor provides sufficient force feedback to safely complete the task.
\section{Conclusion}
This work establishes that reliable manipulation of small, thin objects requires more than an anthropomorphic form; it demands the ability to actively shape and sense contact at the fingertip. The ARISTO Hand addresses this challenge through three integrated capabilities: a rigid sensorized fingernail for edge engagement, a compliant tactile pad for stable acquisition, and active distal hyperextension to align the fingertip with the environment.

Our experiments quantitatively validate this hybrid approach. Active distal hyperextension improved pull-out force on thin objects (1--20~mm) by 2.76$\times$ compared to standard flexion while retaining 98\% of the nominal pull-out capability at credit-card thickness. Additionally, our sensing experiments revealed a fundamental limitation of proprioception in fine manipulation: as fingers straighten into poking or prying configurations, external contact forces become weakly observable through joint torque estimation. By integrating a distal force--torque sensor, ARISTO maintains reliable force observability even near kinematic singularities.

The SD card extraction and insertion task further demonstrates the complementary roles of the sensing and mechanical subsystems. The rigid nail enables recessed edge interaction, while the hyperextending tactile pad stabilizes thin-object acquisition and insertion.

Future work will extend these system-level evaluations to task-level ablation studies that isolate the roles of distal hyperextension, rigid-nail interaction, tactile slip sensing, and proprioceptive estimation across fine-grained manipulation behaviors. We additionally aim to develop richer force-feedback controllers and investigate lower-dimensional sensing configurations for scalable multi-finger implementations.

Taken together, these results suggest that reliable fine-grained manipulation depends not only on dexterous kinematics, but on the ability to deliberately shape and sense contact at the fingertip itself.

\printbibliography


\end{document}